\documentclass[runningheads]{llncs}

\usepackage[T1]{fontenc}
\usepackage{graphicx}
\usepackage{amsmath}
\usepackage{amsmath, amssymb, amsfonts}
\usepackage{booktabs} 
\usepackage{makecell} 
\usepackage{svg}
\usepackage{bm} 
\usepackage{mathtools}
\usepackage{pifont} 
\usepackage{adjustbox}

\usepackage{marvosym}

\usepackage[
    colorlinks=true, 
    citecolor=blue,  
    linkcolor=red,  
    urlcolor=blue   
]{hyperref}

\begin{document}

\title{MINT: Memory-Infused Prompt Tuning at Test-time for CLIP}

\author{Jiaming Yi\inst{1} \and Ruirui Pan\inst{1} \and Jishen Yang\inst{2} \and Xiulong Yang\inst{1}\textsuperscript{(\Letter)}}

\authorrunning{J. Yi et al.}

\institute{School of Computer, Central China Normal University, Wuhan, China \\
\email{\{yijm,rrpan\}@mails.ccnu.edu.cn,yangxiulong@ccnu.edu.cn}
\and Amazon, Seattle, USA \\
\email{jshyang@amazon.com}}

\maketitle

\begin{abstract}
Improving the generalization ability of Vision-Language Pre-trained Models (VLMs) under test-time data distribution shifts remains a critical challenge. The existing Test-Time Adaptation (TTA) methods fall short in fully leveraging the model's internal knowledge, particularly in dynamically adapting to complex and hierarchical visual semantic information. In this paper, we propose \textbf{M}emory-\textbf{In}fused Prompt \textbf{T}uning (MINT), a novel framework to address this issue. Inspired by human associative memory theory, MINT introduces a Memory Prompt Bank (MPB), which stores learnable key-value prompt pairs that work as a memory of previously seen samples. During the test time, relevant prompt pairs in the MPB are retrieved by the hierarchical visual features of test images to dynamically assemble Associative Prompts. The associative prompts are then injected into the image encoder for fine-grained, customized visual contextual guidance. MINT also utilizes learnable text prompts. MINT thus enables rapid, precise VLM adaptation at test time by leveraging this MPB-acquired memory, without source data or retraining.
The code is available at \url{https://github.com/Jamieyi2004/MINT}.

\end{abstract}

\section{Introduction}
Vision-Language Pre-trained Models  such as CLIP~\cite{clip} are foundational for computer vision understanding and generalization, excelling in zero-shot and few-shot learning. Trained on vast image-text data, VLMs encode diverse visual concepts and transfer effectively to various downstream tasks~\cite{task1,task2,task5,task6,task7,task8,task9}. However, the performance of VLMs suffers in Out-of-Distribution (OOD) scenarios-such as style transfers, image degradations, or environmental changes—where test data distributions differ significantly from those of the training data. This issue can be solved by Test-Time Adaptation (TTA) techniques, which adapt models dynamically to new test inputs without access to source training data or full retraining.

Among TTA strategies, prompt tuning is a particularly efficient and flexible solution~\cite{ptnb}. Initially, VLMs employed manually designed text prompts (e.g., “a photo of a [class]”) for downstream tasks.
Although practically applicable, manual prompt design is time-consuming, requires expertise, and lacks generality. To overcome these limitations, subsequent research introduced prompt tuning, treating prompts as learnable parameters optimized with data to better adapt to specific tasks or distributions. In TTA, learnable prompt techniques, such as TPT's~\cite{tpt} text prompt optimization or VPA's~\cite{vpa} visual prompts, help mitigate performance degradation under distribution shifts.

% In Test-Time Adaptation (TTA), learnable prompt techniques, such as Text Prompt Tuning (TPT) [] for text prompt optimization or Visual Prompt Adaptation (VPA) [] for visual prompts, help mitigate performance degradation under distribution shifts.

\begin{figure}[t!]
    \centering 
    \includegraphics[width=0.6\textwidth]{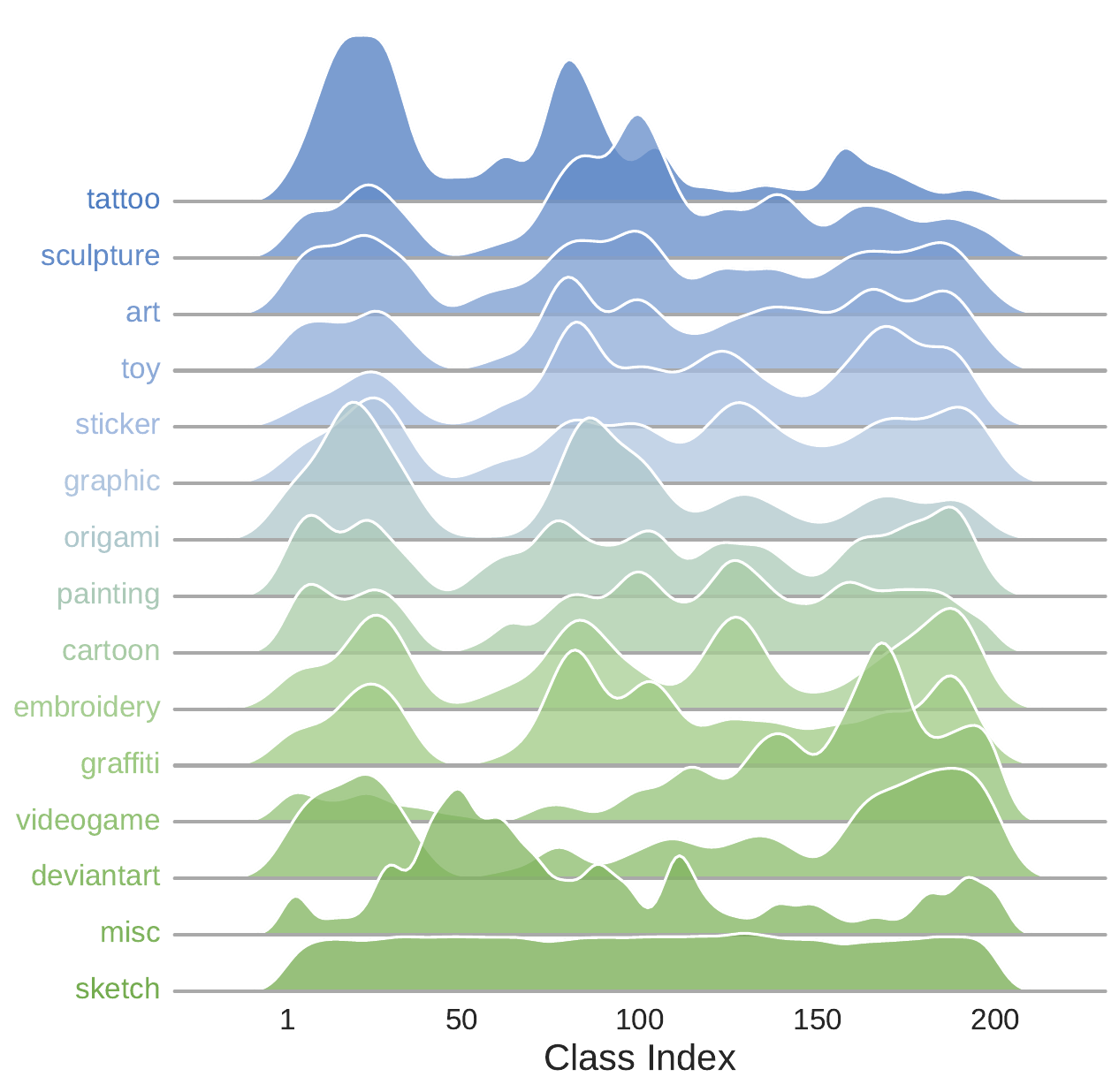} 
    \caption{Distribution of test-time samples on ImageNet-R. Different classes present distinct domain shifts, motivating adaptive strategies for TTA.}
    \label{fig:dis2}
\end{figure}

However, these methods are often constrained by the limited representational capacity of fixed or singular prompt structures, failing to fully exploit the rich, multi-level prior knowledge in pre-trained models, especially in the visual modality. Moreover, since test-time data often originate from multiple domains (see Fig.~\ref{fig:dis2}), relying on a single, static prompt limits the model's capacity to capture both the domain-specific variations and cross-domain commonalities. Specifically, with a dynamic prompt mechanism capable of capturing multi-level semantics, models could more effectively leverage their prior knowledge in a dynamic and compositional manner, providing customized, structured, and sample-specific contextual guidance. Thus, more effectively activating internal knowledge and constructing targeted adaptation strategies based on specific visual inputs is a key challenge in improving TTA performance.

% \begin{figure}[htbp] % htbp 是图片位置选项

To address these challenges, this paper proposes MINT, an innovative test-time memory-infused prompt tuning framework. MINT enhances CLIP's adaptability by using learnable text prompts and introducing a Memory Prompt Bank (MPB) of learnable base visual prompt components. Crucially, this MPB is continuously optimized using the incoming test samples, enabling it to capture and reuse patterns observed in previously seen test data. It extracts hierarchical visual features as queries to dynamically retrieve and integrate base components from the Bank, constructing input-relevant Associative Prompts. Inspired by human associative memory, this mechanism dynamically associates the image's multi-level visual representations with Bank components, selectively activating and combining them into tailored, structured composite prompts for specific samples. These customized prompts capture subtle visual features and provide fine-grained visual contextual guidance, enabling CLIP models to efficiently leverage internal pre-trained knowledge and adapt quickly and accurately to new samples without source data or retraining. This dynamic retrieval and composition process, fueled by the MPB's learning from past test samples, enables the model to strike a balance between adapting to novel domain-specific knowledge and retaining useful information from previously seen domains, which is essential for robust performance in multi-domain test-time scenarios. MINT aims to significantly enhance CLIP's generalization and robustness in complex OOD scenarios. The main contributions of this paper can be summarized as follows:
\begin{itemize}
    \item We propose a novel test-time memory-infused prompt tuning framework (MINT), which constructs and utilizes a dynamic Memory Prompt Bank to more effectively synergize visual and language knowledge for adaptation at test time.
    \item Inspired by human associative memory, our mechanism uses hierarchical semantic features to dynamically combine base prompts from the Memory Prompt Bank, generating customized, fine-grained visual context for each sample and enhancing adaptation precision.
    \item Extensive experiments on multiple benchmark datasets demonstrate the superiority of MINT in significantly enhancing the OOD generalization ability and robustness to test perturbations of CLIP models.
\end{itemize}

\section{Related Work}
\subsection{Prompting for Pre-trained Models}
Pre-trained Vision-Language Models like CLIP~\cite{clip} and ALIGN~\cite{align} exhibit strong zero-shot capabilities with knowledge transferable to diverse downstream tasks. Recent work has proposed methods to effectively transfer this knowledge. Prompting offers a direct way to apply models to downstream tasks in a zero-shot manner. However, hand-crafted prompts are task-sensitive and inefficient.

Recent work has aimed to resolve this by introducing prompt tuning, where prompts are learned directly using training data from downstream tasks. Similar to fine-tuning model parameters, prompts can be optimized using training data because their embeddings are part of the model's input and are differentiable with respect to the loss function. This methodology often yields prompts superior to hand-crafted ones. For instance, CoOp~\cite{coop} and CoCoOp~\cite{cocoop} introduced learnable text prompts, while VPT~\cite{vpt} pioneered learnable visual prompts. Additionally, MaPLe~\cite{maple} proposes to train cross-modal prompts for the adaptation of CLIP. These methods optimize prompts during training with labeled data for specific downstream tasks, typically learning a single or a set of holistic, continuous prompt vectors.

MINT focuses on enhancing the generalization capabilities of VLMs during the test phase, without relying on any additional training or source data. It achieves this by employing a novel memory mechanism and leveraging multi-level semantic information from the visual encoder to dynamically generate adaptive prompts, thereby addressing out-of-distribution challenges.

% TPT的贡献 尤其是实验证明为什么是prompt
\subsection{Test-Time Adaptation}
Test-Time Adaptation (TTA) aims to enable pre-trained models, post-deployment, to dynamically adapt to unseen test data distributions without accessing the original training data. 

Existing TTA methods largely fall into two categories. One class of approaches utilizes self-supervised signals, such as minimizing prediction entropy or imposing consistency regularization through data augmentation, to optimize model parameters or a subset thereof. Another category focuses on adjusting parts of the model's parameters, for instance, by updating Batch Normalization (BN) ~\cite{tent} statistics for test domain characteristics or by modifying the feature extractor while freezing the prediction module. Zhang et al. ~\cite{memo} also showed that optimizing the entire model at test time is feasible. 

While effective in certain scenarios, these methods often involve updating model weights directly, which can lead to overfitting or forgetting, especially under limited test samples. To mitigate this, recent approaches explore \textit{parameter-efficient TTA}. For example, TPT~\cite{tpt} adapts only text prompts during inference, showing that prompt-level updates can preserve pre-trained knowledge while improving test-time robustness.

Despite these advances, existing prompt-based TTA methods typically employ a fixed prompt structure, which limits their expressiveness and adaptability. They also lack mechanisms for memory or reuse of past adaptation experience. In contrast, our method introduces a dynamic, memory-augmented prompt generation strategy that supports sample-specific and compositional prompt synthesis, enhancing both flexibility and generalization under domain shifts.

\begin{figure}[t!]
    \centering % 居中显示
    \includegraphics[width=1\textwidth, trim=4cm 2.2cm 4cm 2.3cm, clip]{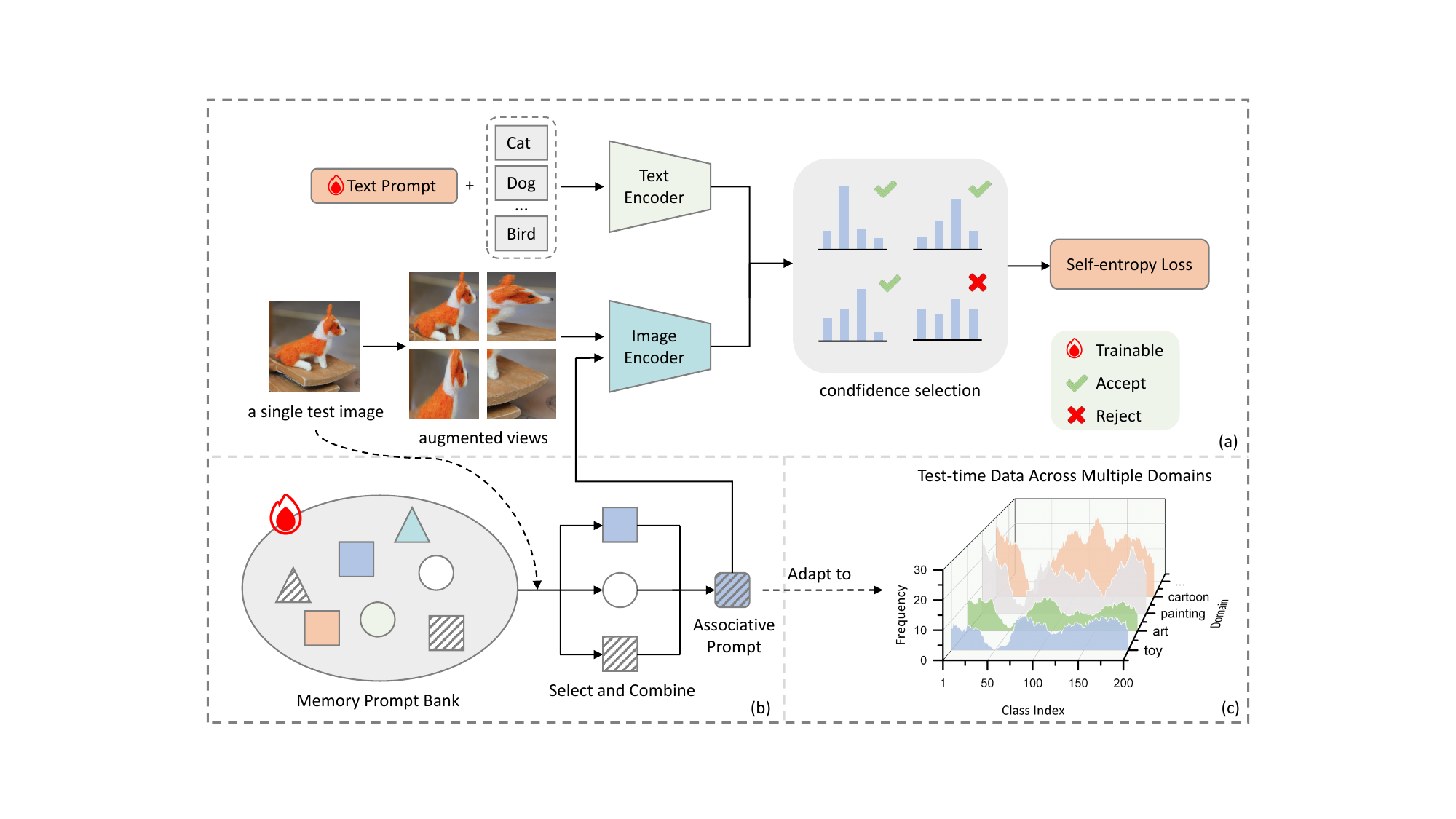} 
    \caption{Overview of our proposed \textbf{M}emory-\textbf{In}fused Prompt \textbf{T}uning (MINT) framework at Test-time for CLIP.}
    \label{fig:mint_overview}
\end{figure}

\section{Method}
We introduce MINT, a Memory-Infused Prompt Tuning framework for test-time adaptation of CLIP. MINT dynamically generates visual prompts by querying a learnable memory bank with hierarchical image features and combines them with learnable text prompts for joint adaptation. This section first reviews the CLIP architecture and conventional prompt tuning, then details the MINT architecture, and finally explains how MINT operates under the test-time adaptation setting.
Fig.~\ref{fig:mint_overview} illustrates the overall MINT framework.

\subsection{Preliminaries: CLIP and Prompt Tuning}

A pre-trained CLIP model, denoted as $\theta = \{E_I, E_T\}$, comprises an image encoder $E_I$ and a text encoder $E_T$. These encoders project images and text descriptions into a shared multimodal embedding space. Typically, $E_I$ is a Vision Transformer (ViT) ~\cite{vit} or a Convolutional Neural Network (CNN) ~\cite{resnet}, while $E_T$ is a Transformer-based architecture ~\cite{transformer}. The model is pre-trained on a vast dataset of (image, text) pairs (which we refer to as $\mathcal{D}_{\text{source}}$), using a contrastive loss function designed to maximize the cosine similarity between embeddings of matched pairs and minimize it for unmatched pairs.

In a standard zero-shot inference setting for a $K$-class classification task, an input image $\mathbf{x}$ is first encoded into a visual feature $\mathbf{v} = E_I(\mathbf{x})$. For the text, a \textit{hand-crafted prompt} is typically used, which is a fixed template, such as "a photo of a", denoted as $P_{\text{h}}$. For each class $c_k$ (where $k=1, \dots, K$), the template is combined with the class name to form a complete textual description, $T_k = [P_{\text{h}}; c_k]$. The descriptions are then fed into the text encoder to obtain class-specific text features: $\mathbf{t}_k = E_T(T_k)$. The prediction probability for class $c_k$ given the image $\mathbf{x}$ is subsequently computed as:
\begin{equation}
\label{eq:clip_prediction}
P(y = c_k | \mathbf{x}) = \frac{\exp(\text{cos}(\mathbf{v}, \mathbf{t}_k)/\tau)}{\sum_{j=1}^{K} \exp(\text{cos}(\mathbf{v}, \mathbf{t}_j)/\tau)},
\end{equation}
where $\tau$ is a temperature hyperparameter.
prompt tuning methods that learn the prompt itself are proposed.
To improve adaptability beyond fixed hand-crafted prompts like $P_{\text{h}}$, \textit{prompt tuning} methods was introduced, where the prompt itself is learned. The goal is to train a prompt that maximizes the performance on a specific downstream task for which labeled data is available. Instead of a fixed $P_{\text{h}}$, a sequence of $L$ learnable prompt vectors, $\mathbf{P}_{\text{l}} = \{\mathbf{p}_1, \mathbf{p}_2, \dots, \mathbf{p}_L\}$, where each $\mathbf{p}_i \in \mathbb{R}^{D_T}$ (where $D_T$ is the text embedding dimension), is optimized. These learnable prompts are typically prepended to the class name embeddings. Thus, for each class $c_k$ in the set of $K$ classes $\mathcal{C}=\{c_1, \dots, c_K\}$, the modified text input becomes $T'_k = [\mathbf{P}_{\text{l}}; c_k]$. The text inputs are then encoded by the text encoder $E_T$ to obtain the prompted text features $\mathbf{t}'_k = E_T(T'_k)$. The learnable prompt $\mathbf{P}_{\text{l}}$ is optimized by minimizing a task-specific loss on a downstream task dataset $\mathcal{D}_{\text{task}} = \{(\mathbf{x}_i, y_i)\}$:
\begin{equation}
\label{eq:prompt_tuning_objective} % Ensure this label is consistent if referenced
\mathbf{P}_{\text{l}}^* = \arg\min_{\mathbf{P}_{\text{l}}} \mathbb{E}_{(\mathbf{x}, y) \sim \mathcal{D}_{\text{task}}} \mathcal{L}_{\text{task}}(f(\mathbf{x}, [\mathbf{P}_{\text{l}}; \mathcal{C}]), y),
\end{equation}
where $f(\mathbf{x}, [\mathbf{P}_{\text{l}}; \mathcal{C}])$ represents the CLIP prediction process. The process involves obtaining the visual feature $\mathbf{v} = E_I(\mathbf{x})$ (using the image encoder $E_I$) and the set of prompted text features $\{\mathbf{t}'_1, \dots, \mathbf{t}'_K\}$ (derived from $[\mathbf{P}_{\text{l}}; c_k]$ for each $c_k \in \mathcal{C}$ using the text encoder $E_T$). The prediction probabilities are then typically computed similarly to Equation~\eqref{eq:clip_prediction}, but using $\mathbf{v}$ and these prompted text features $\mathbf{t}'_k$. $\mathcal{L}_{\text{task}}$ is the task-specific loss, commonly a cross-entropy loss, calculated based on these probabilities.

\subsection{Memory-Infused Prompt Architecture and Generation}
\label{subsec:memory_infused_prompts_arch_gen}

MINT uses two core prompt mechanisms, both comprising learnable components that are dynamically infused with information and optimized at test-time: (i) learnable text prompts and (ii) dynamically generated Associative Prompts derived from a Memory Prompt Bank (MPB).

First, \textbf{Learnable Text Prompts} are utilized to adapt the textual input to the target domain. A set of $L_t$ learnable tokens $\mathbf{P}_{\text{t}} = \{\mathbf{p}_{\text{t},j}\}_{j=1}^{L_t}$ are prepended to each class name embedding $E_{\text{class}}$ to form $[\mathbf{P}_{\text{t}}; E_{\text{class}}(c_k)]$ for the text encoder $E_T$ (as in Fig.~\ref{fig:mint_overview}a). The tokens $\mathbf{P}_{\text{t}}$ are updated during test-time adaptation.

Second, adaptive visual guidance is synthesized via a \textbf{Memory Prompt Bank}, $\mathbf{MPB}$ (see Fig.~\ref{fig:mint_overview}b). The MPB stores $N_{\text{MPB}}$ learnable (key, value) pairs:
\begin{equation}
\label{eq:mpb_definition}
\mathbf{MPB} = \{(\mathbf{k}_i, \mathbf{v}_i)\}_{i=1}^{N_{\text{MPB}}},
\end{equation}
where $\mathbf{k}_i \in \mathbb{R}^{D_I}$ denotes the $i$-th retrieval key, and $\mathbf{v}_i \in \mathbb{R}^{L_m \times D_I}$ represents the $i$-th Memory Prompt. $D_I$ is the dimensionality of the image encoder tokens, and $L_m$ signifies the number of constituent tokens within each Memory Prompt $\mathbf{v}_i$.
The elements of both $\mathbf{k}_i$ and $\mathbf{v}_i$ are initialized by sampling from a standard normal distribution, $\mathcal{N}(0,1)$.
Both the set of keys $\{\mathbf{k}_i\}_{i=1}^{N_{\text{MPB}}}$ and the set of Memory Prompts $\{\mathbf{v}_i\}_{i=1}^{N_{\text{MPB}}}$ are learnable parameters. Multiple such Memory Prompts are subsequently combined to form an Associative Prompt.

The constituent keys $\{\mathbf{k}_i\}$ and Memory Prompts $\{\mathbf{v}_i\}$ of the MPB are dynamically updated during test-time adaptation. These Memory Prompts serve as building blocks for Associative Prompts injected into the image encoder.
 
To construct the dynamic Associative Prompt $\mathbf{P}_{\text{a}}$ for a test image $\mathbf{x}$, MINT performs:

% TODO
\noindent\textbf{Image-based Querying:}
For a given image $\mathbf{x}$, query features $\{\mathbf{q}^{(l)}\}_{l=1}^{N_{\text{layers}}}$ are extracted from $N_{\text{layers}}$ distinct levels of the image encoder $E_I$.
Let the output of the $l$-th layer of the encoder, $E_I^{(l)}(\mathbf{x})$, be a sequence of $N_I$ tokens:
$$ E_I^{(l)}(\mathbf{x}) = [\mathbf{z}_{0}^{(l)}, \mathbf{z}_{1}^{(l)}, \dots, \mathbf{z}_{N_{I-1}}^{(l)}] $$
where each token $\mathbf{z}_{i}^{(l)} \in \mathbb{R}^{D_I}$. The query feature $\mathbf{q}^{(l)}$ is defined as the first token in this sequence (often the [CLS] token equivalent for ViTs if $i=0$ is used for that purpose, or simply the first patch token):
$$ \mathbf{q}^{(l)} = \mathbf{z}_{0}^{(l)} $$
Each such query $\mathbf{q}^{(l)} \in \mathbb{R}^{D_I}$ encapsulates a distinct level of visual abstraction from the corresponding layer.

\noindent\textbf{Memory Retrieval:}
For each query $\mathbf{q}^{(l)}$, cosine similarity with every key $\mathbf{k}_i$ in the Memory Prompt Bank is computed:
\begin{equation}
\label{eq:similarity_score}
S(\mathbf{q}^{(l)}, \mathbf{k}_i) = \frac{\mathbf{q}^{(l)} \cdot \mathbf{k}_i}{\|\mathbf{q}^{(l)}\|_2 \cdot \|\mathbf{k}_i\|_2}.
\end{equation}

The top $N_{\text{sel}}$ entries for query $\mathbf{q}^{(l)}$ are selected based on similarity. Let $\mathcal{J}^{(l)}_{\text{sel}}$ be the set of indices of these entries:
\begin{equation}
\label{eq:top_k_selection}
\mathcal{J}^{(l)}_{\text{sel}} = \operatorname*{arg\,topK}_{i \in \{1, \dots, N_{\text{MPB}}\}}^{N_{\text{sel}}} S(\mathbf{q}^{(l)}, \mathbf{k}_i).
\end{equation}
yielding a set of Memory Prompts $\{\mathbf{v}_j \mid j \in \mathcal{J}^{(l)}_{\text{sel}}\}$ most relevant to $\mathbf{q}^{(l)}$.

\noindent\textbf{Prompt Composition:}
Memory Prompts from all layers are aggregated to form an Associative Prompt:

\begin{equation}
\label{eq:prompt_composition}
\mathbf{P}_{\text{a}} = \frac{1}{|\mathcal{V}_{\text{sel}}|} \sum_{\mathbf{v} \in \mathcal{V}_{\text{sel}}} \mathbf{v}, \quad \text{where } \mathcal{V}_{\text{sel}} = \bigcup_{l=1}^{N_{\text{layers}}} \{\mathbf{v}_j \mid j \in \mathcal{J}^{(l)}_{\text{sel}}\}.
\end{equation}
Here, $\mathbf{P}_{\text{a}} \in \mathbb{R}^{L_m \times D_I}$, where averaging is performed if all selected Memory Prompts $\mathbf{v}_m$ share compatible dimensions; otherwise, alignment or projection is used to ensure consistency.

\noindent\textbf{Associative Prompt Injection:}
This consolidated Associative Prompt $\mathbf{P}_{\text{a}}$ is subsequently injected into the image encoder $E_I$. In the MINT framework, this injection occurs exclusively at the input to the first layer of $E_I$. For Vision Transformer (ViT)~\cite{vit} architectures, if $\mathbf{E}_{\text{patch}}(\mathbf{x})$ represents the sequence of embedded image patches, the input to the first Transformer block becomes $[\mathbf{P}_{\text{a}}; \mathbf{E}_{\text{patch}}(\mathbf{x})]$, effectively prepending the dynamic Associative Prompt.

\subsection{Test-Time Adaptation (TTA) with MINT}

MINT extends CLIP and prompt tuning into a fully test-time adaptation (FTTA) setting, where neither the original pre-training data $\mathcal{D}_{\text{source}}$ nor ground-truth labels $\{y_i\}$ for the target domain $\mathcal{D}_{\text{task}}$ are accessible during the adaptation phase. Thus, while Equation~\eqref{eq:prompt_tuning_objective} describes supervised prompt tuning, MINT operates solely on unlabeled test images $\{\mathbf{x}_i\}$.

During TTA, MINT adapts to new test data online—often in batches with augmented views (see Fig.~\ref{fig:mint_overview}a)—without requiring source data or model retraining. The learnable parameters, denoted $\theta_{\text{MINT}}$, include the text prompts $\mathbf{P}_{\text{t}}$ and the key-value pairs $\{\mathbf{k}_i, \mathbf{v}_i\}$ within the MPB.

For each incoming test image $\mathbf{x}_i$, dynamic visual prompts $\mathbf{P}_{\text{a},i}$ are generated by querying the MPB with hierarchical features extracted from $\mathbf{x}_i$. The Associative Prompts are injected into the image encoder $E_I$ to obtain adapted visual features $\mathbf{v}'_i$. Simultaneously, the learnable text prompt $\mathbf{P}_{\text{t}}$ is combined with class names to produce textual features $\{\mathbf{t}_k\}$ (see Fig.~\ref{fig:mint_overview}). The prediction probabilities are then computed as:

\[
P(y|\mathbf{x}_i) = \text{softmax}(\cos(\mathbf{v}'_i, \mathbf{t}_k)/\tau).
\]

A confidence-based selection step (see Fig.~\ref{fig:mint_overview}a) is applied to filter uncertain predictions. MINT then optimizes the following loss function to adjust its learnable components:

\begin{equation}
\label{eq:tta_loss_condensed}
\mathcal{L} = \mathbb{E}_{\mathbf{x}_i \sim \text{batch}} \left[ H(P(y|\mathbf{x}_i)) 
- \lambda \sum_{l=1}^{N_{\text{layers}}} \! \!
\sum_{\mathbf{k} \in \mathcal{K}^{(l)}_{\text{sel}}}
S(\mathbf{q}^{(l)}, \mathbf{k}) \right]
,
\end{equation}

where $H(p) = - \sum_k p_k \log p_k$ and $\mathcal{K}^{(l)}_{\text{sel}} = \{\mathbf{k}_j \mid j \in \mathcal{J}^{(l)}_{\text{sel}}\}$. $\lambda$ is a weighting parameter. The learnable prompts $\theta_{\text{MINT}}$ are then updated via gradient descent to adapt the prompt strategies continuously across varying test-time distributions.

Unlike methods that modify model weights directly or batch normalization statistics, MINT preserves the pre-trained model's parameters and adapts solely via lightweight, semantically grounded prompts. The associative memory mechanism of MINT supports compositional reuse of prompt components, allowing both sample-specific adaptation and robust performance across different domains present in the test data.

\section{Experiments}
In this section, we present the experimental evaluation conducted for MINT. We first outline the benchmarks for assessment and comparison with leading methods, followed by implementation specifics. We conclude with comprehensive results, analyses, and an ablation study of our proposed method.
\subsection{Datasets}
To evaluate model performance against common real-world distribution shifts, we adopt the setup that Radford et al.~\cite{clip} proposed. Robustness is assessed on four ImageNet~\cite{i} variants, detailed subsequently. These are established out-of-distribution (OOD) benchmarks, providing a rigorous test for generalization.
\begin{itemize}
    
    \item \textbf{ImageNet-R}~\cite{ir} includes 30,000 images across 200 ImageNet categories, characterized by diverse artistic renditions (e.g., art, cartoons, paintings) that introduce substantial domain variations.
    \item \textbf{ImageNet-A}~\cite{ia}, comprising 7,500 natural adversarial images across 200 ImageNet categories. These capture challenging variations, forming multiple 'failure mode domains' and introducing significant distribution shifts.
    \item \textbf{ImageNet-V2}~\cite{iv2} serves as an independent test set, containing 10,000 natural images across the 1,000 ImageNet categories, collected from different sources than the original validation data.
    \item \textbf{ImageNet-Sketch}~\cite{ik} consists of 50,000 black-and-white sketch images, with 50 examples for each of the 1,000 ImageNet classes, offering a distinct stylistic deviation.

\end{itemize}

\subsection{Baselines}
To evaluate our method, we compare it against representative baselines from zero-shot inference, few-shot fine-tuning, and test-time adaptation.
\subsubsection{Zero-shot Inference.}
For zero-shot, standard CLIP is directly applied to downstream tasks with a fixed prompt "a photo of a [class]". Additionally, an Ensemble strategy~\cite{clip} aggregates predictions from 80 hand-crafted prompts to improve generalization.
\subsubsection{Few-shot Fine-tuning Baselines.}
These baselines use 16 labeled samples per category for optimization. CoOp~\cite{coop} learns continuous, dataset-specific text prompt vectors. CoCoOp~\cite{cocoop} generates dynamic, input-conditional prompts via a meta-network using visual features. MaPLe~\cite{maple} learns multi-modal prompts for both text and visual encoders, enabling joint cross-modal optimization.
\subsubsection{Test-Time Adaptation Baselines.}
Test-time adaptation (TTA) methods adjust model parameters online using unlabeled test data for domain shifts. TPT~\cite{tpt} performs online prompt tuning for CLIP with unlabeled test samples. SAR~\cite{sar} enhances TTA robustness via selective parameter updates based on confidence and consistency, mitigating error accumulation. ZERO~\cite{zero} leverages N-fold augmented views and zero-temperature softmax for efficient, backpropagation-free adaptation. MTA~\cite{mta} employs a MeanShift approach with jointly optimized 'inlierness scores' for augmented views, offering training-free adaptation.

\subsection{Implementation Details} 

MINT is implemented using the pre-trained CLIP model with a ViT-B/16 backbone.
The Memory Prompt Bank (MPB) holds $N_{\text{MPB}} = 512$ learnable memory prompt components. Each memory prompt component $\mathbf{v}_i$ within the MPB consists of $L_m=2$ tokens. For each query $\mathbf{q}^{(l)}$, we retrieve the top $N_{\text{sel}}=3$ memory prompt components from the MPB based on cosine similarity with the keys $\{\mathbf{k}_i\}$. The aggregated Associative Prompt $\mathbf{P}_{\text{a}}$ is then added to the first layer of the image encoder. During test-time adaptation, each test image generates $B=64$ augmented views via random resized cropping. The confidence-based strategy selects the top $\kappa=10\%$ most confident (lowest self-entropy) samples for loss calculation. The hyperparameter $\lambda$ balancing the entropy and similarity terms in the loss is set to $0.2$. Optimization is performed using the AdamW optimizer with a learning rate of $5 \times 10^{-3}$. All experiments are implemented on a single NVIDIA A100 GPU.

\begin{table}[t!]

\centering
\footnotesize
\caption{Top-1 accuracy comparison results on various ImageNet benchmarks.}
\label{table1}

\begin{tabular}{@{}l cc cc cc c@{}} 
\toprule 
Methods         & Publication   &ImageNet-R       & ImageNet-A        & ImageNet-V2           & ImageNet-Sketch & \thead{Average} \\
\midrule       
CLIP            & ICML'21       &73.98            & 47.87             & 60.86                 & 46.09           & 57.20 \\
Ensemble        & ICML'21       &77.65            & 49.89             & 61.88                 & 48.24           & 59.42 \\
\midrule                                                             
CoOp            & IJCV'22       &75.21            & 49.71             & 64.20                 & 47.99           & 59.28 \\
CoCoOp          & CVPR'22       &76.18            & 50.63             & 64.07                 & 48.75           & 59.91 \\
MaPLe           & CVPR'23       &76.98            & 50.90             & 64.07                 &\underline{49.15}& 60.28 \\
\midrule                                                             
TPT             & NeurIPS'22       &77.06            & 54.77             & 63.45                 & 47.94           & 60.81 \\
SAR             & ICLR'23       &75.81            & 48.89             & 63.29                 & 47.59           & 58.90 \\ 
MTA             & CVPR'24       &\underline{78.33}& 58.06             & \underline{64.24}     & \textbf{49.61}  & 62.56 \\ 
ZERO            & NeurIPS'24       &77.28            & \textbf{61.35}    & 64.13                 & 48.29           & \underline{62.76} \\ 
\midrule   
\textbf{MINT}   & \textbf{Ours} &\textbf{78.68}   & \underline{59.83} & \textbf{65.83}        & 48.16           & \textbf{63.12} \\  

\bottomrule
\end{tabular}
% } 
\end{table}

\subsection{Results}
To validate the effectiveness of our proposed MINT model, we compare it against the aforementioned representative baselines, including zero-shot inference, few-shot fine-tuning, and test-time adaptation (TTA). The detailed Top-1 accuracy results of these comparisons are presented in Table ~\ref{table1}.

MINT achieves an average Top-1 accuracy of 63.12\% across the four datasets, securing the top rank and significantly outperforming all compared models, including the leading few-shot fine-tuning method MaPLe (60.28\%) and the top-performing test-time adaptation method ZERO (62.76\%). This superior average performance provides initial evidence of MINT's generalization capability and robustness in complex out-of-distribution (OOD) scenarios. Notably, MINT attains an optimal accuracy of 78.68\% on ImageNet-R, a dataset characterized by diverse artistic styles that introduce substantial domain variations. This strongly supports MINT's objective to significantly enhance CLIP's generalization and robustness in complex OOD scenarios.

In summary, MINT not only leads in average performance but also achieves a breakthrough on ImageNet-R, a critical benchmark for multi-domain adaptation and knowledge retention. These findings fully demonstrate its superiority and design effectiveness in complex OOD testing scenarios.

% 组件消融
\begin{table}[htbp]
\centering
\footnotesize 
\caption{Ablation study on the effect of different components on ImageNet-A. The \checkmark signifies that the respective component is included.}
\label{table2}

{ 
\setlength{\tabcolsep}{6pt} % 将列间距减半
\begin{tabular}{@{}cccc@{}}

\toprule
% 使用 \thead 使长标题可以换行，减少列宽
\thead{Associative \\ Prompt} & \thead{General \\ Visual Prompt} & \thead{Learnable \\ Text Prompt} & \thead{Top-1 \\ Accuracy}  \\
\midrule
           & \checkmark &                 & 49.12      \\ % 有普通视觉提示 没有文本提示
\checkmark &            &                 & 49.25      \\ % 有联想提示 没有文本提示
           &            & \checkmark      & 55.16  \\ % 只有文本提示 
           & \checkmark & \checkmark      & \underline{58.03}      \\ % 普通视觉提示 有文本提示 
\checkmark &            & \checkmark      & \textbf{59.83}  \\ % ours
\bottomrule
\end{tabular}
} 
\end{table}

\subsection{Ablation Study}
\subsubsection{The Effect of Different Components.}
We conduct ablation studies (Table~\ref{table2}) to evaluate the contribution of MINT's core components. A General Visual Prompt $\mathbf{P}_{\text{v}}$ is a standard learnable visual prompt, distinct from our dynamic Associative Prompt $\mathbf{P}_{\text{a}}$ generated via the Memory Prompt Bank. The Learnable Text Prompt $\mathbf{P}_{\text{t}}$ alone yields 55.16\% accuracy. Combining $\mathbf{P}_{\text{t}}$ with $\mathbf{P}_{\text{v}}$ improves this to 58.03\%. Notably, the full MINT configuration ($\mathbf{P}_{\text{a}} + \mathbf{P}_{\text{t}}$) achieves the highest accuracy at 59.83\%. This demonstrates the superiority of the MINT dynamic Associative Prompt over a $\mathbf{P}_{\text{v}}$ and validates our design. 

\subsubsection{The Effect of Different Hyperparameters.}
\begin{figure}[]
    \centering % 居中显示
    \includegraphics[width=\textwidth, trim=1.5cm 0cm 1.5cm 1cm, clip]{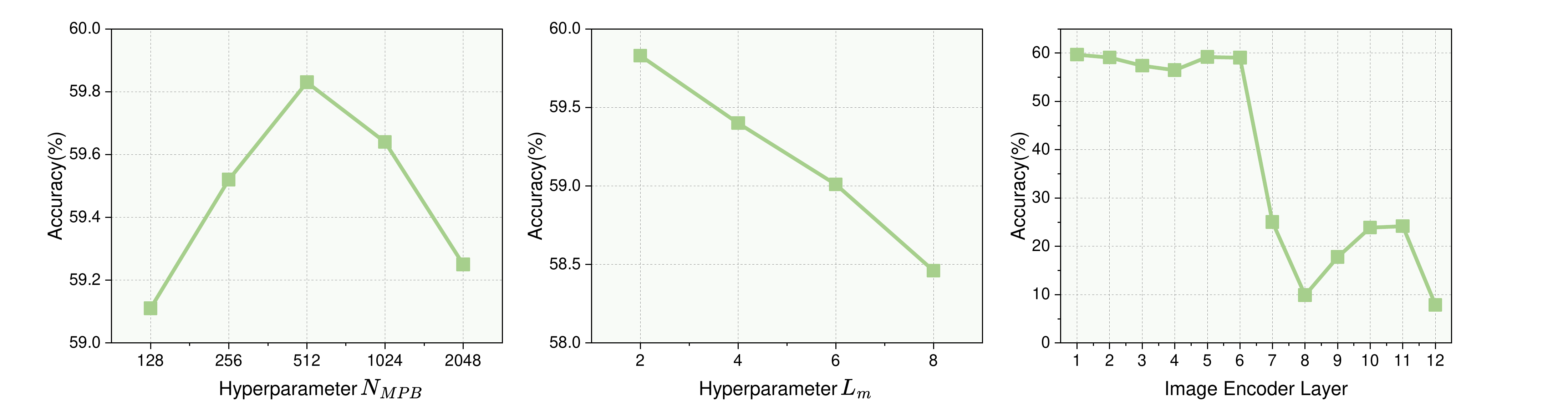} 
    \caption{Ablation study about the effect of different hyper-parameters on ImageNet-A.}
    \label{fig:hp}
\end{figure}

We further conduct ablation studies (Fig.~\ref{fig:hp}) to investigate the influence of MINT's key hyperparameters. First, we investigated $N_{\text{MPB}}$, the total number of learnable entries in the Memory Prompt Bank. Optimal performance was observed with $N_{\text{MPB}}=512$. This MPB size provides sufficient capacity to store diverse visual-semantic patterns, crucial for robust adaptation, while maintaining a manageable complexity for effective learning and retrieval. We then examined $L_m$, which defines the length of each Memory Prompt component $\mathbf{v}_i$. Optimal performance was achieved when $L_m=2$. This setting, under limited test data, effectively balances token expressiveness with the parameter complexity for adaptation.  Lastly, we investigated the image encoder layer for adding the Associative Prompt $\mathbf{P}_{\text{a}}$. Considering that only one sample is available for immediate adaptation at test time, to ensure the stability and efficiency of model adaptation, we add prompts to only a single layer. Experiments show that adding to the first layer performs best. 

\section{Conclusion}
This paper introduces MINT, an innovative test-time framework enhancing VLM (e.g., CLIP) generalization under distribution shifts. Its core memory mechanism, inspired by human associative memory, builds a Memory Prompt Bank (MPB) of learnable base components. Hierarchical visual features query this MPB to dynamically generate and integrate customized Associative Prompts injected into the image encoder for fine-grained visual guidance, synergizing with learnable text prompts. Experiments show MINT significantly enhances CLIP's OOD generalization and robustness. This stems from its leveraging not only internal pre-trained knowledge but, critically, the memory of test-time data itself. MINT's MPB effectively learns commonalities and specificities among test samples, thereby transcending reliance on static knowledge or weight modification.

\textbf{Limitations.} MINT's dynamic MPB mechanisms introduce computational overhead, especially for large-scale or high-frequency test data. The initialization and hyperparameters of MPB components may require scenario-specific tuning.

\textbf{Future Directions.} MINT's memory-driven dynamic prompting opens several future research avenues. Exploring more efficient memory retrieval and composition to reduce costs and improve speed is key. Extending MINT to other models and multimodal tasks (e.g., video understanding, VQA) could validate its versatility. Overall, MINT offers a valuable pathway for enhancing the adaptability of large pre-trained models in dynamic settings.

\bibliographystyle{splncs04} 
\bibliography{main} 

\end{document}